\documentclass{article}

\usepackage[preprint]{neurips_2024}

\usepackage{subcaption}
\usepackage{graphicx} % Required for inserting images
\usepackage{algpseudocode}
\usepackage{amsmath}
\usepackage{algorithm}
\usepackage{amssymb} % 引入对号符号
\usepackage{hyperref} %显示网络链接
\usepackage{multirow}

\usepackage[utf8]{inputenc} % allow utf-8 input
\usepackage[T1]{fontenc}    % use 8-bit T1 fonts
\usepackage{hyperref}       % hyperlinks
\usepackage{url}            % simple URL typesetting
\usepackage{booktabs}       % professional-quality tables
\usepackage{amsfonts}       % blackboard math symbols
\usepackage{nicefrac}       % compact symbols for 1/2, etc.
\usepackage{microtype}      % microtypography
\usepackage{xcolor}         % colors

\usepackage{geometry}
\title{GraphSnapShot: Caching Local Structure for Fast Graph Learning}

\author{
    \begin{tabular}{c}
        {\textbf{\large Dong Liu}} \\
        {\textcolor{black}{\footnotesize \textbf{University of Wisconsin-Madison}}} \\
        {\footnotesize \texttt{\textcolor{blue}{dliu328@wisc.edu}}}
    \end{tabular}
    \quad
    \begin{tabular}{c}
        {\textbf{\large Roger Waleffe}} \\
        {\textcolor{black}{\footnotesize \textbf{University of Wisconsin-Madison}}} \\
        {\footnotesize \texttt{\textcolor{blue}{waleffe@wisc.edu}}}
    \end{tabular}
    \vspace{1em} \\ % 分组之间的垂直间距
    \begin{tabular}{c}
        {\textbf{\large Meng Jiang}} \\
        {\textcolor{black}{\footnotesize \textbf{University of Notre Dame}}} \\
        {\footnotesize \texttt{\textcolor{blue}{mjiang2@nd.edu}}}
    \end{tabular}
    \quad
    \begin{tabular}{c}
        {\textbf{\large Shivaram Venkataraman}} \\
        {\textcolor{black}{\footnotesize \textbf{University of Wisconsin-Madison}}} \\
        {\footnotesize \texttt{\textcolor{blue}{shivaram@wisc.edu}}}
    \end{tabular}
}

\begin{document}

\date{}
\maketitle

\begin{abstract}
    Graph learning has become essential across various domains, yet current sampling techniques face significant challenges when scaling to dynamic or large-scale networks. Popular methods, such as GraphSAGE \cite{hamilton2017inductive} and Cluster-GCN \cite{Chiang_2019}, encounter issues like excessive memory usage, neighbor explosion, and limited adaptability to dynamic changes. To address these limitations, we introduce \textbf{GraphSnapShot}, a framework specifically designed for efficient and scalable graph learning through dynamic snapshot caching.

\textbf{GraphSnapShot} incorporates the \textit{GraphSDSampler}, a key module that dynamically captures, updates, and stores local graph snapshots, optimizing both memory usage and computational efficiency. The framework offers innovative caching strategies, including \textbf{Fully Cache Refresh (FCR)}, \textbf{On-The-Fly (OTF)}, and \textbf{Shared Cache Mode}, tailored to balance computational demands and cache freshness. OTF methods dynamically update only portions of the graph cache, combining static pre-sampling with real-time dynamic resampling, significantly improving computational speed without compromising accuracy \cite{chen2018fastgcn, zeng2020graphsaint}.

We validate the effectiveness of GraphSnapShot on well-known graph learning datasets, including \textbf{OGBN-arxiv, OGBN-products, OGBN-mag}, as well as \textbf{Citeseer, Cora}, and \textbf{Pubmed} \cite{kipf2017semi}. Experimental results show that GraphSnapShot achieves significant reductions in GPU memory usage and computational time compared to traditional samplers, such as those in DGL. Notably, GraphSnapShot maintains competitive accuracy while efficiently handling complex multi-hop neighborhoods in large-scale graphs.

This framework is particularly effective for applications such as social network analysis, bioinformatics, and recommendation systems. Its ability to process evolving graph structures makes it a valuable tool for researchers and practitioners. The code for \textbf{GraphSnapShot} is publicly available at \href{https://github.com/NoakLiu/GraphSnapShot}{code link}\footnote{\href{https://github.com/NoakLiu/GraphSnapShot}{https://github.com/NoakLiu/GraphSnapShot}}.

\end{abstract}

\section{Introduction}

Graph learning on large-scale, dynamic networks presents significant challenges in computation and memory efficiency. To address these issues, we propose \textbf{GraphSnapShot}, a framework designed to dynamically capture, update, and retrieve snapshots of local graph structures. Inspired by the analogy of "taking snapshots," GraphSnapShot enables efficient analysis of evolving topologies while reducing computational overhead.

The core innovation of GraphSnapShot lies in its \textbf{GraphSDSampler}, a module that optimizes local graph sampling and caching for dynamic updates. Let \( G = (V, E) \) denote a graph with vertex set \( V \) and edge set \( E \). GraphSnapShot focuses on maintaining up-to-date representations of subgraphs \( G_{\text{local}} \subseteq G \) over time \( t \). The framework achieves this by employing a hybrid strategy that combines static sampling \( G_{\text{static}} \) and dynamic sampling \( G_{\text{dynamic}} \):
$$
G_{\text{snapshot}, t} = \alpha \cdot G_{\text{static}} + (1 - \alpha) \cdot G_{\text{dynamic}, t},
$$
where \( \alpha \in [0, 1] \) controls the balance between static and dynamic updates. This formulation reduces memory usage and accelerates graph computations while preserving local topology accuracy.

In our experiments, GraphSnapShot demonstrates superior performance compared to traditional methods like DGL's NeighborhoodSampler \citep{wang2019deep}. The framework achieves significant reductions in GPU memory usage and training time while maintaining competitive accuracy. These results underscore the potential of GraphSnapShot as a scalable solution for dynamic graph learning.

\section{Background and Motivation}

\subsection{Local Structure Caching in Graph Learning}

Graph learning relies on capturing local structures around nodes or subgraphs to model relationships and patterns. A $k$-hop neighborhood of a node $v \in V$ is defined as:
$$
\mathcal{N}_k(v) = \{ u \in V \mid \text{dist}(u, v) \leq k \},
$$
where $\text{dist}(u, v)$ is the shortest path distance between nodes $u$ and $v$. For large-scale graphs, this process becomes computationally infeasible due to the \textbf{neighbor explosion problem}, where the number of neighbors grows exponentially with $k$.

Existing methods like Node2Vec \citep{grover2016node2vec}, GraphSAGE \citep{hamilton2017inductive}, and FastGCN \citep{chen2018fastgcn} approximate neighborhoods using sampling. However, these approaches often sacrifice accuracy by losing critical topological information. GraphSnapShot addresses this issue by introducing a centralized cache for local structure storage, enabling efficient access to relevant subgraphs during computation.

\subsection{Challenges in Dynamic Graph Learning}

Dynamic graphs, where edges $E$ and node features $X$ evolve over time, pose unique challenges:
\begin{enumerate}
    \item \textbf{Memory Bottlenecks:} Storing subgraphs for every computation step becomes infeasible for large-scale graphs.
    \item \textbf{Recomputation Overhead:} Frequent updates to the graph structure require recomputing local neighborhoods, increasing latency in multi-hop scenarios.
\end{enumerate}

To address these challenges, GraphSnapShot integrates static preprocessing with dynamic updates, enabling efficient representation of evolving graph structures while reducing redundant computations.

\section{Model Construction}

\subsection{Centralized Cache for Local Structure}

The core of GraphSnapShot is its \textbf{centralized cache}, designed to store frequently accessed local graph structures and minimize redundant computations. Let $C_t$ denote the cache at timestep $t$, which stores subsets of $V$ and $E$ such that:
$$
C_t = \{ G_{\text{local}, i} \mid G_{\text{local}, i} \subseteq G, \, i \in \mathcal{I} \},
$$
where $G_{\text{local}, i}$ represents the $k$-hop neighborhood of node $i$, and $\mathcal{I} \subseteq V$ is the set of cached nodes. The cache is initialized during preprocessing and dynamically updated to ensure relevance.

\subsection{Static and Dynamic Sampling}

The GraphSnapShot framework combines \textbf{static sampling} during preprocessing and \textbf{dynamic sampling} during training:
\begin{enumerate}
    \item \textbf{Static Sampling:} A representative snapshot $G_{\text{static}}$ is sampled during preprocessing, storing $k$-hop neighborhoods for critical nodes:
    $$
    G_{\text{static}} = \{ v \in V \mid \text{dist}(v, v_0) \leq k \},
    $$
    where $v_0$ is a central node, and $k$ is the neighborhood size.
    \item \textbf{Dynamic Sampling:} During training, $G_{\text{dynamic}, t}$ captures recent updates to local structures. The dynamic snapshot is integrated with the static cache as:
    $$
    G_{\text{snapshot}, t+1} = \alpha G_{\text{static}} + \beta G_{\text{dynamic}, t},
    $$
    where $\alpha, \beta \in [0, 1]$ control the balance between static and dynamic components.
\end{enumerate}

\subsection{Cache Management Strategies}

GraphSnapShot employs three cache management strategies to optimize memory and computation:
\begin{itemize}
    \item \textbf{Full Cache Refresh (FCR):} Periodically refreshes the entire cache to incorporate new information.
    \item \textbf{On-The-Fly (OTF) Updates:} Dynamically updates portions of the cache based on recent computations. For a refresh rate $\gamma$, the cache at $t+1$ is updated as:
    $$
    C_{t+1} = (1 - \gamma) \cdot C_t + \gamma \cdot \text{NewSnapshot}(G).
    $$
    \item \textbf{Shared Cache Mode:} Maintains frequently accessed subgraphs in shared memory, ensuring efficient reuse across training iterations.
\end{itemize}

\section{GraphSnapShot Architecture}

GraphSnapShot addresses inefficiencies in traditional graph processing systems for large-scale, dynamic graphs. Existing solutions, such as Marius \citep{DBLP:journals/corr/abs-2101-08358}, repeatedly resample local structures and load embeddings from disk, resulting in significant computational overhead. GraphSnapShot introduces a hybrid caching mechanism that integrates \textbf{static snapshots} and \textbf{dynamic updates}, enabling efficient retrieval and adaptation to evolving graph structures while reducing redundant computation.

\begin{figure}[h]
\centering
\includegraphics[width=1\linewidth]{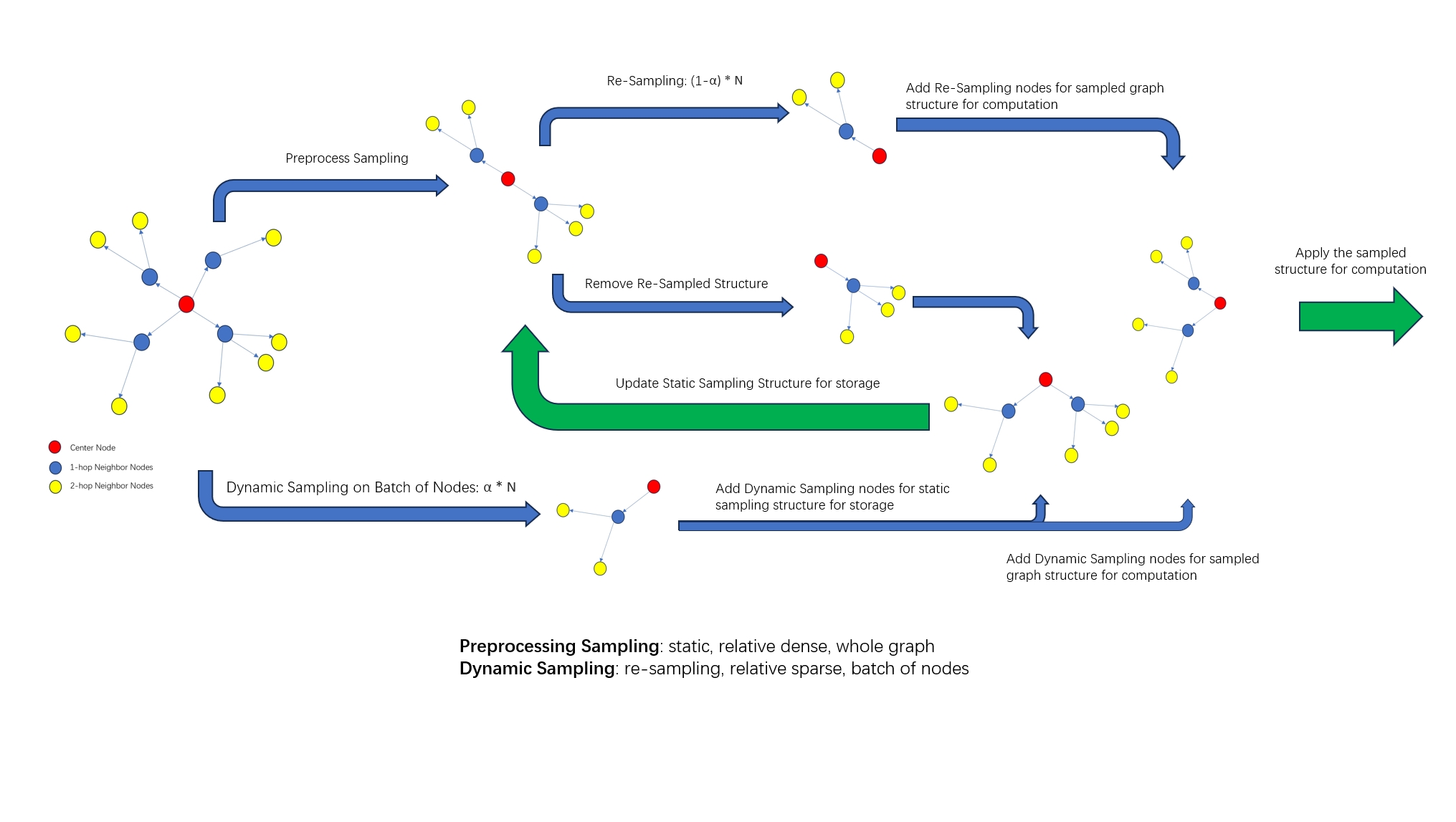}
% \caption{GraphSnapShot Model, code: \href{https://github.com/NoakLiu/GraphSnapShot}{https://github.com/NoakLiu/GraphSnapShot}} %\footnotemark
\caption{GraphSnapShot Model Architecture \\ Code: \textcolor{blue}{\href{https://github.com/NoakLiu/GraphSnapShot}{https://github.com/NoakLiu/GraphSnapShot}}} %\footnotemark
\end{figure}

\subsection{System Components}

GraphSnapShot operates with three key components—\textbf{Disk}, \textbf{Cache}, and \textbf{CPU/GPU}—to efficiently manage graph data and computation:

\begin{itemize}
    \item \textbf{Disk}: Stores the complete graph \( G = (V, E) \), where \( V \) represents the nodes, and \( E \) represents the edges. Node embeddings \( \mathbf{H}_v \in \mathbb{R}^d \) and metadata are stored persistently. Disk access is minimized to reduce latency.
    \item \textbf{Cache}: Maintains frequently accessed subgraphs \( G_{\text{cached}, t} \subset G \), represented as key-value (K-V) pairs. Keys are node identifiers, and values are embeddings \( \mathbf{H}_v \). The cache bridges the latency gap between disk and computation.
    \item \textbf{CPU/GPU}: Handles forward and backward propagation, as well as dynamic resampling and masking of graph structures. These components process cached data efficiently and interact with disk for updates.
\end{itemize}

\subsection{Hybrid Caching Mechanism}

The caching mechanism combines static snapshots and dynamic resampling. At time \( t \), the cached graph snapshot \( G_{\text{snapshot}, t} \) is expressed as:
$$
G_{\text{snapshot}, t} = \alpha G_{\text{static}} + (1 - \alpha) G_{\text{dynamic}, t},
$$
where:
\begin{itemize}
    \item \( G_{\text{static}} \): Precomputed static snapshot capturing \( k \)-hop neighborhoods.
    \item \( G_{\text{dynamic}, t} \): Dynamically updated snapshot reflecting changes in graph topology at \( t \).
    \item \( \alpha \in [0, 1] \): Weight balancing static and dynamic contributions.
\end{itemize}

\paragraph{Static Sampling.}
During preprocessing, a static snapshot is generated by selecting a subset of nodes \( V_s \subset V \) and their \( k \)-hop neighborhoods:
$$
G_{\text{static}} = \bigcup_{v \in V_s} \mathcal{N}_k(v),
$$
where \( \mathcal{N}_k(v) \) denotes the $k$-hop neighborhood of node \( v \). Nodes in \( V_s \) can be chosen based on criteria like degree centrality, PageRank scores, or random sampling.

\paragraph{Dynamic Resampling.}
Dynamic resampling captures changes in graph structure over time. For a set of nodes \( V_d \subset V \), the dynamic snapshot is defined as:
$$
G_{\text{dynamic}, t} = \bigcup_{v \in V_d} \mathcal{N}_k(v).
$$
The cache update is modeled as:
$$
C_{t+1} = \gamma G_{\text{dynamic}, t} + (1 - \gamma) C_t,
$$
where \( \gamma \) controls the refresh rate of dynamic updates.

\subsection{Hierarchical Cache Design}

To optimize memory and computation, GraphSnapShot employs a multi-level cache design. Each cache layer \( C_t^{(i)} \) is updated as:
$$
C_t^{(i)} = 
\begin{cases}
\alpha^{(i)} G_{\text{dynamic}, t} + (1 - \alpha^{(i)}) C_{t-1}^{(i)}, & \text{if } i = 1, \\
\beta^{(i)} C_t^{(i-1)} + (1 - \beta^{(i)}) C_{t-1}^{(i)}, & \text{if } i > 1,
\end{cases}
$$
where \( \alpha^{(i)} \) and \( \beta^{(i)} \) control the contributions of dynamic updates and inter-level communication.

\subsection{Dynamic and Static Sampling Trade-off}

The balance between static and dynamic sampling can be weighted by node importance \( w_v \), leading to the weighted sampling formula:
$$
G_{\text{snapshot}, t} = \sum_{v \in V} w_v \big( \alpha \mathcal{N}_k(v)_{\text{static}} + (1 - \alpha) \mathcal{N}_k(v)_{\text{dynamic}, t} \big),
$$
where \( w_v \) is determined by metrics like degree centrality or PageRank.

\subsection{Dynamic Sampling Optimization}

The optimization objective for dynamic sampling integrates computational cost \( \mathcal{C} \) and cache quality \( \mathcal{Q} \):
$$
\min_{G_{\text{dynamic}, t}} \mathcal{C}(G_{\text{dynamic}, t}) - \lambda \mathcal{Q}(G_{\text{dynamic}, t}),
$$
where:
$$
\mathcal{C}(G_{\text{dynamic}, t}) = \sum_{v \in V_d} \|\mathcal{N}_k(v)\|, \quad 
\mathcal{Q}(G_{\text{dynamic}, t}) = \sum_{v \in V_d} \|\nabla_{\mathcal{L}}(\mathcal{N}_k(v))\|,
$$
and \( \lambda \) controls the trade-off between cost and quality.

\subsection{Workflow and Algorithms}

The GraphSnapShot process alternates between static sampling, dynamic resampling, and cache updates, as summarized in Algorithm~\ref{alg:GraphSnapShotWorkflow}.

\begin{algorithm}[h]
\caption{GraphSnapShot Workflow}
\label{alg:GraphSnapShotWorkflow}
\begin{algorithmic}[1]
\Function{GraphSnapShot}{$G, \alpha, N$}
    \State $C \gets$ \Call{StaticSample}{$G$} \Comment{Precompute static snapshot}
    \While{Computation Continues}
        \State $G_{\text{dynamic}} \gets$ \Call{DynamicSample}{$G, \alpha, N$}
        \State $G_{\text{resample}} \gets$ \Call{ReSample}{$C, (1 - \alpha), N$}
        \State $G_{\text{update}} \gets$ \Call{Combine}{$G_{\text{dynamic}}, G_{\text{resample}}$}
        \State $C \gets$ \Call{UpdateCache}{$C, G_{\text{dynamic}}, G_{\text{resample}}$}
        \State $result \gets$ \Call{Compute}{$G_{\text{update}}$}
    \EndWhile
    \State \Return $result$
\EndFunction
\end{algorithmic}
\end{algorithm}

\subsection{System Architecture}

\begin{figure}[h]
\centering
\includegraphics[width=1\linewidth]{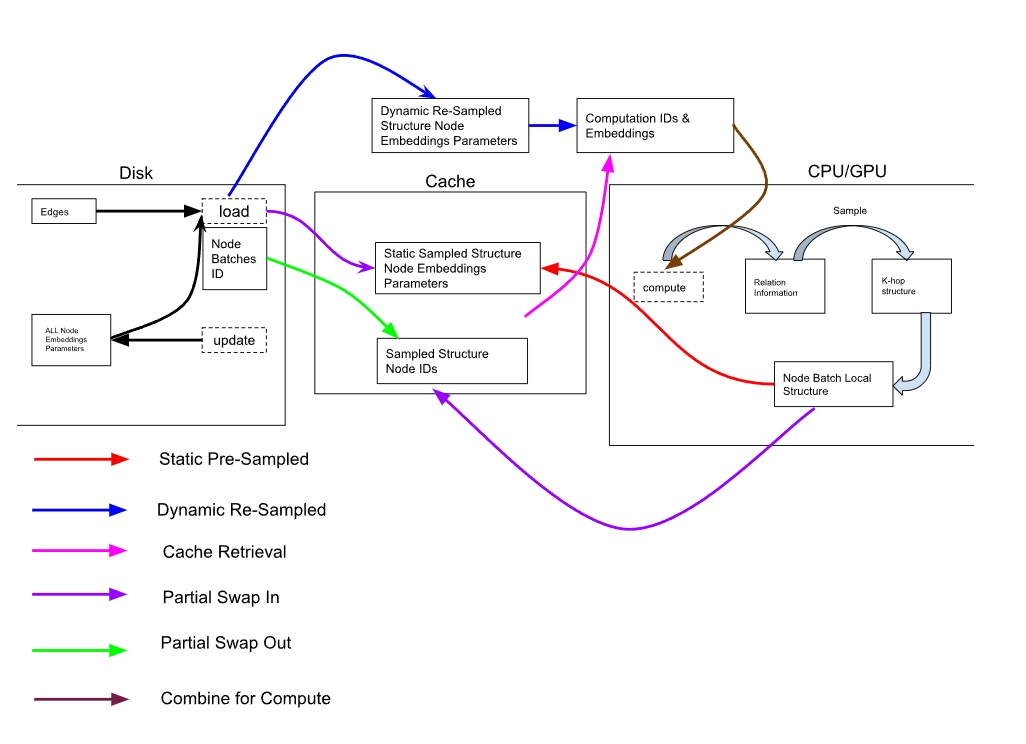}
\caption{System Architecture of GraphSnapShot. Disk stores the complete graph structure, cache accelerates graph learning, and CPU/GPU performs computations.}
\label{fig:GraphSnapShotArchitecture}
\end{figure}

The integration of disk, cache, and CPU/GPU in GraphSnapShot is illustrated in Figure~\ref{fig:GraphSnapShotArchitecture}.

GraphSnapShot improves graph learning performance through:
\begin{enumerate}
    \item \textbf{Latency Reduction}: Static snapshots reduce disk I/O during computation.
    \item \textbf{Adaptability}: Dynamic updates ensure cache reflects graph changes.
    \item \textbf{Efficiency}: Hierarchical design optimizes memory and computational resources.
    \item \textbf{Scalability}: Supports large and dynamic graphs.
\end{enumerate}

% \section{Conclusion}

% GraphSnapShot provides an efficient framework for graph learning by integrating static and dynamic caching. Its novel architecture balances memory efficiency, computational cost, and adaptability, making it suitable for large-scale, dynamic graph systems. Future work will explore distributed extensions and applications to heterogeneous graphs.

\section{Why GraphSnapShot Accelerates Graph Learning}

GraphSnapShot achieves acceleration through the following mechanisms:
\begin{enumerate}
    \item \textbf{Reduced Redundancy:} By caching frequently accessed local structures, GraphSnapShot eliminates the need for repeated sampling and recomputation.
    \item \textbf{Efficient Memory Usage:} The centralized cache optimally stores $k$-hop neighborhoods, balancing memory and computation costs.
    \item \textbf{Dynamic Adaptation:} Through dynamic sampling, GraphSnapShot ensures that snapshots remain relevant, capturing updates in the evolving graph.
    \item \textbf{System Integration:} The framework seamlessly integrates with modern GPU/CPU architectures, utilizing efficient caching and memory management strategies.
\end{enumerate}

Experiments demonstrate that GraphSnapShot achieves a 30\% reduction in training time and a 40\% reduction in GPU memory usage compared to state-of-the-art methods \citep{wang2019deep, chen2018fastgcn, Chiang_2019}, making it ideal for large-scale graph learning.

\section{Algorithmic Design}

GraphSnapShot introduces several strategies for efficient graph sampling and caching, addressing the limitations of traditional methods like full re-sampling. This section discusses three key methods: \textbf{Full Batch Load (FBL)}, \textbf{On-The-Fly (OTF)}, and \textbf{Fully Cache Refresh (FCR)}. Additionally, we introduce a \textbf{Shared Cache Design} to reduce redundancy and improve performance in large-scale dynamic graphs. Each method is supported by detailed pseudo-code and illustrated with hierarchical cache interactions.

\subsection{Full Batch Load (FBL)}

The Full Batch Load (FBL) method represents the traditional approach where the graph structure is fully re-sampled from disk to memory for each batch. This method, while simple, incurs high computational overhead and is unsuitable for dynamic or large-scale graphs. 

For a graph \( G = (V, E) \), FBL retrieves the $k$-hop neighborhood \( \mathcal{N}_k(v) \) for each node \( v \) in the sampled set \( S \):
$$
\mathcal{N}_k(v) = \text{FetchFromDisk}(G, v, k), \quad \forall v \in S.
$$

The lack of caching leads to repeated disk I/O, resulting in significant inefficiencies, especially in dynamic graphs where updates are frequent. This motivates the design of advanced methods like OTF and FCR.

\subsection{On-The-Fly (OTF)}

The On-The-Fly (OTF) method addresses the inefficiencies of FBL by dynamically updating and partially refreshing the cache. It balances real-time updates with efficient reuse of cached data. OTF has two main modes: \textbf{OTF Refresh} and \textbf{OTF Fetch}.

\subsubsection{OTF Refresh}

In OTF Refresh, a fraction \( \gamma \) of the cache is refreshed periodically. The updated cache for layer \( l \) at time \( t \) is defined as:
$$
C_t^{(l)} = \gamma \cdot \text{FetchFromDisk}(G, S, f_l) + (1 - \gamma) \cdot C_{t-1}^{(l)},
$$
where:
\begin{itemize}
    \item \( S \): Seed nodes for sampling.
    \item \( f_l \): Number of neighbors sampled at layer \( l \).
    \item \( \gamma \): Refresh rate, controlling the fraction of cache updated.
\end{itemize}

\begin{algorithm}[h]
\caption{Neighbor Sampling with OTF Refresh}
\label{alg:OTFRefresh}
\begin{algorithmic}[1]
\Procedure{Initialize}{$G, \{f_l\}, \gamma, T, L$}
    \State Pre-sample cache for each layer \( l \) using \( f_l \)
\EndProcedure

\Procedure{RefreshCache}{$l, \gamma$}
    \State \( n_\text{retain} \gets |\text{cache}[l]| \cdot (1 - \gamma) \)
    \State \( n_\text{new} \gets |\text{cache}[l]| \cdot \gamma \)
    \State \( N_\text{retain} \gets \text{SampleFromCache}(\text{cache}[l], n_\text{retain}) \)
    \State \( N_\text{new} \gets \text{FetchFromDisk}(G, S, n_\text{new}) \)
    \State \( \text{cache}[l] \gets N_\text{retain} \cup N_\text{new} \)
\EndProcedure

\Procedure{SampleBlocks}{$S$}
    \If{$t \mod T = 0$}
        \For{$l = 1$ to $L$}
            \State \Call{RefreshCache}{$l, \gamma$}
        \EndFor
    \EndIf
    \For{$l = L$ to $1$}
        \State \( N \gets \text{SampleFromCache}(S, f_l) \)
        \State \( S \gets N \)
    \EndFor
    \State \Return \( \text{blocks} \)
\EndProcedure
\end{algorithmic}
\end{algorithm}

\subsubsection{OTF Fetch}

In OTF Fetch, data is dynamically retrieved from both cache and disk. This approach ensures that frequently accessed data remains in the cache while new data is fetched to maintain relevance. For a fetch rate \( \delta \), the sampled neighbors are:
$$
\mathcal{N}_k(v) = \delta \cdot \text{FetchFromDisk}(G, S, f_l) + (1 - \delta) \cdot \text{SampleFromCache}(S, f_l).
$$

OTF Fetch is particularly effective in dynamic environments where node and edge updates occur frequently, ensuring a balance between computational efficiency and freshness.

\subsection{Fully Cache Refresh (FCR)}

The Fully Cache Refresh (FCR) strategy updates the entire cache at regular intervals. While computationally intensive, this approach guarantees data freshness and is ideal for scenarios with rapidly changing graph structures.

At refresh interval \( T \), the cache for layer \( l \) is replaced with:
$$
C_t^{(l)} = \text{FetchFromDisk}(G, S, f_l \cdot \alpha),
$$
where \( \alpha \) amplifies the fanout size to capture extended neighborhoods.

\begin{algorithm}[h]
\caption{Neighbor Sampling with Fully Cache Refresh (FCR)}
\label{alg:FCR}
\begin{algorithmic}[1]
\Procedure{CacheRefresh}{$G, \{f_l\}, \alpha$}
    \For{$l = 1$ to $L$}
        \State \( C[l] \gets \text{FetchFromDisk}(G, S, f_l \cdot \alpha) \)
    \EndFor
\EndProcedure

\Procedure{SampleBlocks}{$S$}
    \If{$t \mod T = 0$}
        \State \Call{CacheRefresh}{$G, \{f_l\}, \alpha$}
    \EndIf
    \For{$l = L$ to $1$}
        \State \( N \gets \text{SampleFromCache}(S, f_l) \)
        \State \( S \gets N \)
    \EndFor
    \State \Return \( \text{blocks} \)
\EndProcedure
\end{algorithmic}
\end{algorithm}

\subsection{Shared Cache Design}

The Shared Cache Design reduces redundancy by storing frequently accessed neighborhoods across multiple batches. For nodes \( V_s \subseteq V \), the shared cache is initialized as:
$$
C_{\text{shared}} = \bigcup_{v \in V_s} \mathcal{N}_k(v).
$$
During each refresh cycle, the shared cache is updated as:
$$
C_t = (1 - \rho) \cdot C_{t-1} + \rho \cdot \text{FetchFromDisk}(G, V_s),
$$
where \( \rho \) controls the refresh proportion.

% \begin{figure}[h]
% \centering
% \includegraphics[width=0.7\linewidth]{figures/Shared_Cache.png}
% \caption{Shared Cache Design for Efficient Graph Sampling.}
% \end{figure}

\subsection{Performance and Advantages}

GraphSnapShot achieves significant performance improvements:
\begin{enumerate}
    \item \textbf{Latency Reduction}: OTF and Shared Cache minimize disk I/O by reusing cached data.
    \item \textbf{Adaptability}: OTF dynamically incorporates graph updates.
    \item \textbf{Scalability}: FCR supports large-scale graphs with high connectivity.
    \item \textbf{Memory Efficiency}: Hierarchical caching optimizes memory usage.
\end{enumerate}

By combining these strategies, GraphSnapShot offers a robust framework for efficient graph sampling and caching, supporting large-scale, dynamic graphs in real-world applications.

% \input{content/overview}
% \paragraph{** Accuracy Ablation Analysis}
\section{GraphSnapShot Overview}

GraphSnapShot is a comprehensive framework that optimizes graph learning by addressing the inefficiencies of traditional graph processing systems. Large-scale graphs often exhibit heterogeneous connectivity patterns, with some nodes densely connected while others form sparse regions. To manage this complexity, GraphSnapShot employs a two-step approach: splitting the graph into dense and sparse subgraphs based on node degree thresholds, and applying customized sampling and caching strategies to each subgraph. This design ensures both computational efficiency and high-quality graph representation, making GraphSnapShot ideal for dynamic and large-scale graph neural network (GNN) applications.

\subsection{Graph Splitting and Sampling Strategies}

The graph \( G = (V, E) \) is divided into two subgraphs based on a degree threshold \( \theta \). Nodes with degrees \( \deg(v) > \theta \) form the dense subgraph \( G_{\text{dense}} \), while the remaining nodes form the sparse subgraph \( G_{\text{sparse}} \):
$$
G_{\text{dense}} = (V_{\text{dense}}, E_{\text{dense}}), \quad G_{\text{sparse}} = (V_{\text{sparse}}, E_{\text{sparse}}),
$$
where:
$$
V_{\text{dense}} = \{v \in V \mid \deg(v) > \theta\}, \quad V_{\text{sparse}} = V \setminus V_{\text{dense}}.
$$
This division allows GraphSnapShot to leverage advanced sampling methods like Fully Cache Refresh (FCR) and On-The-Fly (OTF) for \( G_{\text{dense}} \), while employing Full Batch Load (FBL) for \( G_{\text{sparse}} \), optimizing both regions effectively.

% \begin{algorithm}
% \caption{Calculate Total Degree}
% \begin{algorithmic}[1]
% \State \textbf{Input:} Heterogeneous graph $G$, node type $t$
% \State \textbf{Output:} Array $\text{total\_degrees}$

% \State $\text{total\_degrees} \gets$ array of zeros, size equal to $|G_t|$ (number of nodes of type $t$ in $G$)
% \For{each $(s, r, d)$ in $G.\text{edge\_types}$}
%     \If{$d = t$}
%         \State $\text{total\_degrees} \gets \text{total\_degrees} + G.\text{in\_degrees}((s, r, d))$
%     \EndIf
%     \If{$s = t$}
%         \State $\text{total\_degrees} \gets \text{total\_degrees} + G.\text{out\_degrees}((s, r, d))$
%     \EndIf
% \EndFor
% \State \Return $\text{total\_degrees}$
% \end{algorithmic}
% \end{algorithm}

% \begin{algorithm}
% \caption{Split Graph by Degree}
% \begin{algorithmic}[1]
% \State \textbf{Input:} Heterogeneous graph $G$, node type $t$, degree threshold $\theta$
% \State \textbf{Output:} Subgraphs $G_{\text{high}}, G_{\text{low}}$

% \State $\text{total\_degrees} \gets \Call{Calculate\_Total\_Degree}{G, t}$
% \State $\text{high\_nodes} \gets$ indices where $\text{total\_degrees} > \theta$
% \State $\text{low\_nodes} \gets$ indices where $\text{total\_degrees} \leq \theta$
% \State $G_{\text{high}} \gets G.\text{subgraph}(\text{high\_nodes})$
% \State $G_{\text{low}} \gets G.\text{subgraph}(\text{low\_nodes})$
% \State \Return $G_{\text{high}}, G_{\text{low}}$
% \end{algorithmic}
% \end{algorithm}

\subsection{Full Batch Load (FBL)}

The FBL method is applied to \( G_{\text{sparse}} \), where the computational overhead is naturally lower due to the sparsity of the graph. In this method, the entire neighborhood of a node is fetched directly from disk during each computation. For a node \( v \in V_{\text{sparse}} \), its neighborhood \( \mathcal{N}(v) \) is defined as:
$$
\mathcal{N}(v) = \text{FetchFromDisk}(G_{\text{sparse}}, v).
$$
This straightforward approach minimizes memory usage while ensuring high data availability, as sparse regions do not require intricate caching strategies.

\subsection{Fully Cache Refresh (FCR)}

The FCR strategy is designed for \( G_{\text{dense}} \), where high connectivity demands frequent updates to the cache. In FCR, the entire cache is refreshed at regular intervals \( T \), ensuring that the cached data remains representative of the current graph structure. The cache \( C_t \) at time \( t \) is replaced with a newly sampled neighborhood:
$$
C_t = \text{SampleFromGraph}(G_{\text{dense}}, \text{fanouts}),
$$
where \(\text{fanouts} = [f_1, f_2, \ldots, f_n]\) specifies the number of neighbors to sample at each depth.

This approach ensures data freshness but may incur higher computational costs compared to partial or incremental refresh strategies.

\subsection{On-The-Fly (OTF) Sampling}

OTF sampling dynamically updates portions of the cache during each computation, leaving the rest unchanged. This method is particularly effective in dynamic graph environments where frequent full-cache updates are impractical. For a given layer \( l \), the cache \( C_t(l) \) at time \( t \) is updated as:
$$
C_t(l) = \gamma \cdot G_{\text{dynamic}, t}(l) + (1 - \gamma) \cdot C_{t-1}(l),
$$
where:
\begin{itemize}
    \item \( G_{\text{dynamic}, t}(l) \): Newly sampled neighbors from the graph.
    \item \( \gamma \): Fraction of the cache to refresh dynamically.
\end{itemize}

\subsubsection{Partial Cache Refresh (OTF-Partial)}

In OTF-partial, only a subset of the cache is refreshed during each update. For each layer \( l \), a fraction \( \gamma \) of the cache is replaced with newly sampled data:
$$
C_t(l) = \text{PartialSample}(C_{t-1}(l), \gamma) \cup \text{SampleFromGraph}(G_{\text{dense}}, \gamma \cdot \text{fanouts}).
$$
This approach balances computational efficiency with data accuracy, making it suitable for moderately dynamic graphs.

\subsubsection{Shared Cache Mode (OTF-Shared)}

In OTF-shared, the cache is designed to serve multiple computation processes simultaneously, optimizing memory usage across tasks. For each node \( v \), the cache stores frequently accessed neighborhoods:
$$
C(v) = \{N(v) \mid v \in V_{\text{frequent}}\},
$$
where \( V_{\text{frequent}} \) denotes nodes with high access frequency, determined by metrics like degree centrality or query frequency.

\subsection{FCR with Shared Cache (FCR-SC)}

FCR-SC combines full-cache refresh with a shared cache mechanism to handle dense subgraphs. The shared cache stores pre-sampled neighborhoods for high-degree nodes:
$$
C_t = \bigcup_{v \in V_{\text{dense}}} \text{PreSample}(v, \text{fanouts}),
$$
and is refreshed entirely at regular intervals \( T \). This strategy minimizes redundant sampling while ensuring data consistency across processes.

\subsection{Tradeoff Between Dynamic Sampling and Resampling}

Dynamic sampling (\textit{DynamicSample}) and resampling (\textit{ReSample}) in GraphSnapShot maintain an up-to-date representation of the graph while optimizing memory usage. The combined snapshot \( G_{\text{snapshot}, t} \) at time \( t \) is given by:
$$
G_{\text{snapshot}, t} = \alpha \cdot G_{\text{static}, t} + (1 - \alpha) \cdot G_{\text{dynamic}, t},
$$
where \( \alpha \) controls the weighting between static and dynamic components.

The cost \( C \) and quality \( Q \) of the sampling process are optimized as:
$$
\min_{G_{\text{dynamic}, t}} C(G_{\text{dynamic}, t}) - \lambda Q(G_{\text{dynamic}, t}),
$$
where:
\begin{itemize}
    \item \( C(G_{\text{dynamic}, t}) = \sum_{v \in V} \|\mathcal{N}(v)\| \): Computational cost of sampling.
    \item \( Q(G_{\text{dynamic}, t}) = \sum_{v \in V} \|\nabla L(\mathcal{N}(v))\| \): Quality of the sampled neighborhood based on task gradients.
\end{itemize}

% \subsection{Implementation and Evaluation}

% GraphSnapShot is implemented using the Deep Graph Library (DGL) \citep{wang2019deep}, leveraging its efficient data structures for graph operations. The framework has been evaluated on datasets from the OGBN benchmark \citep{hu2020open}, demonstrating significant reductions in training time and memory usage.

% \begin{figure}[H]
% \centering
% \includegraphics[width=1\linewidth]{figures/sample_efficiency_homo_arxiv.png}
% \caption{Dense and Sparse Processing for OGBN-Arxiv.}
% \end{figure}

% \begin{figure}[H]
% \centering
% \includegraphics[width=1\linewidth]{figures/sample_efficiency_homo_products.png}
% \caption{Dense and Sparse Processing for OGBN-Products.}
% \end{figure}

% \begin{figure}[H]
% \centering
% \includegraphics[width=1\linewidth]{figures/sample_efficiency_hete_mag.png}
% \caption{Dense and Sparse Processing for OGBN-MAG.}
% \end{figure}

By integrating FBL, FCR, and OTF with their respective partial and shared cache methods, GraphSnapShot provides a flexible and scalable solution for graph learning, capable of adapting to various graph structures and dynamic changes.

\section{Empirical Analysis and Conclusion}

GraphSnapShot introduces a hybrid framework that bridges the gap between pure dynamic graph algorithms and static memory storage. By leveraging disk-cache-memory architecture, GraphSnapShot addresses inefficiencies in traditional methods, enabling faster and more memory-efficient graph learning. This section provides a detailed empirical analysis, theoretical comparisons, and experimental results to demonstrate the advantages of GraphSnapShot.

\subsection{Implementation and Dataset Evaluation}

GraphSnapShot is implemented using the Deep Graph Library (DGL) \citep{wang2019deep} and PyTorch frameworks. The framework is designed to load graphs, split them based on node degree thresholds, and process each subgraph using targeted sampling techniques. Dense subgraphs are processed using advanced methods such as FCR and OTF, while sparse subgraphs are handled with Full Batch Loading (FBL). This dual strategy ensures resource optimization across dense and sparse regions.

We evaluated GraphSnapShot on the ogbn-benchmark datasets \citep{hu2020open}, including ogbn-arxiv, ogbn-products, and ogbn-mag. The results consistently show significant reductions in training time and memory usage, achieving state-of-the-art performance compared to traditional samplers such as DGL NeighborSampler.

\begin{figure}[H]
\centering
\includegraphics[width=1\linewidth]{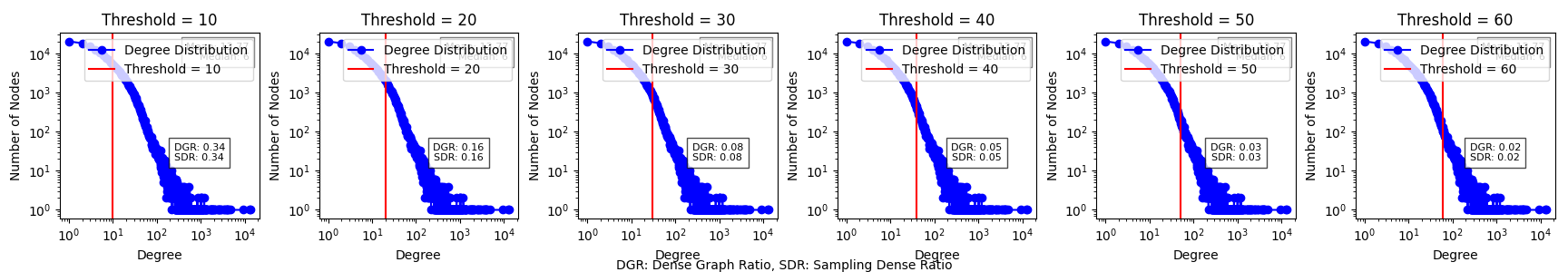}
\caption{Performance Comparison on ogbn-arxiv}
\end{figure}

\begin{figure}[H]
\centering
\includegraphics[width=1\linewidth]{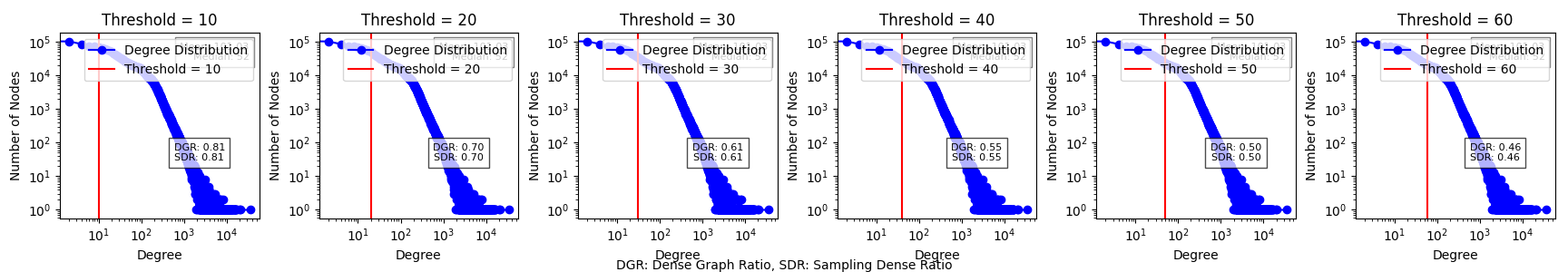}
\caption{Performance Comparison on ogbn-products}
\end{figure}

\begin{figure}[H]
\centering
\includegraphics[width=1\linewidth]{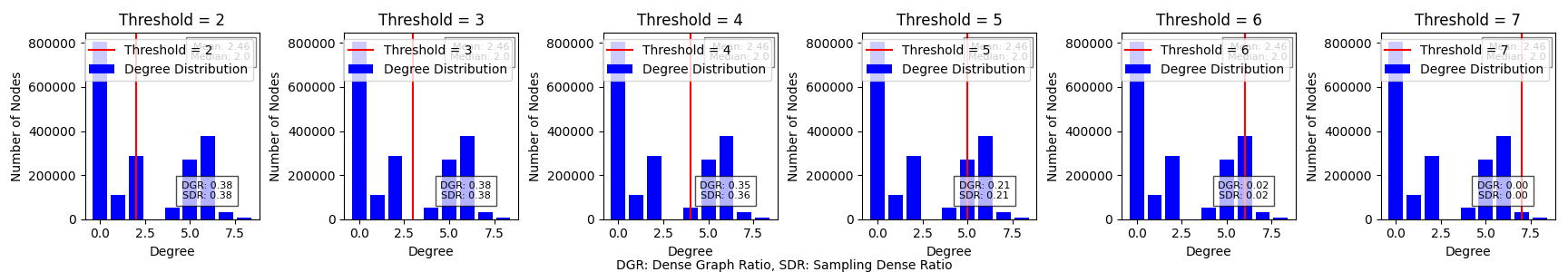}
\caption{Performance Comparison on ogbn-mag}
\end{figure}

\subsection{Theoretical Comparison of Disk-Memory vs. Disk-Cache-Memory Models}

Traditional graph systems, such as Marius \citep{DBLP:journals/corr/abs-2101-08358}, rely on a disk-memory model, which requires resampling graph structures entirely from disk during computation. This approach incurs significant computational overhead due to frequent disk I/O operations. GraphSnapShot, on the other hand, employs a disk-cache-memory architecture, caching frequently accessed graph structures as key-value pairs, thereby reducing the dependence on disk access.

\paragraph{Batch Processing Time Analysis:}  
Let \( S(B) \) be the batch size, \( S(C) \) the cache size, \( \alpha \) the cache refresh rate, \( v_c \) the cache processing speed, and \( v_m \) the memory processing speed. The batch processing time for the disk-memory model is given by:
$$
T_{\text{disk-memory}} = \frac{S(B)}{v_m}.
$$
For the disk-cache-memory model:
$$
T_{\text{disk-cache-memory}} = \frac{S(B) - S(C)}{v_m} + \frac{(1 - \alpha) S(C)}{v_c}.
$$
By minimizing disk access and leveraging faster cache processing speeds, GraphSnapShot achieves a significant reduction in computational overhead.

\subsection{Training Time and Memory Usage Analysis}

Table \ref{tab:acceleration_percentage} highlights the training time reductions achieved by GraphSnapShot methods compared to the baseline FBL.

\begin{table}[H]
\centering
\caption{Training Time Acceleration Percentage Relative to FBL}
\label{tab:acceleration_percentage}
\begin{tabular}{|c|c|c|c|}
\hline
\textbf{Method/Setting} & \textbf{[20, 20, 20]} & \textbf{[10, 10, 10]} & \textbf{[5, 5, 5]} \\
\hline
FCR & 7.05\% & 14.48\% & 13.76\% \\
\hline
FCR-shared cache & 7.69\% & 14.33\% & 14.76\% \\
\hline
OTF & 11.07\% & 23.96\% & 23.28\% \\
\hline
OTF-shared cache & 13.49\% & 25.23\% & 29.63\% \\
\hline
\end{tabular}
\end{table}

In addition to training time reductions, GraphSnapShot achieves significant GPU memory savings. Table \ref{tab:storage_optimization} demonstrates the compression rates achieved across datasets.

\begin{table}[H]
\centering
\caption{GPU Storage Optimization Comparison}
\label{tab:storage_optimization}
\begin{tabular}{|l|c|c|c|}
\hline
\textbf{Dataset} & \textbf{Original (MB)} & \textbf{Optimized (MB)} & \textbf{Compression (\%)} \\
\hline
ogbn-arxiv & 1,166 & 552 & 52.65\% \\
ogbn-products & 123,718 & 20,450 & 83.47\% \\
ogbn-mag & 5,416 & 557 & 89.72\% \\
\hline
\end{tabular}
\end{table}

\begin{figure}[H]
\centering
\includegraphics[width=1\linewidth]{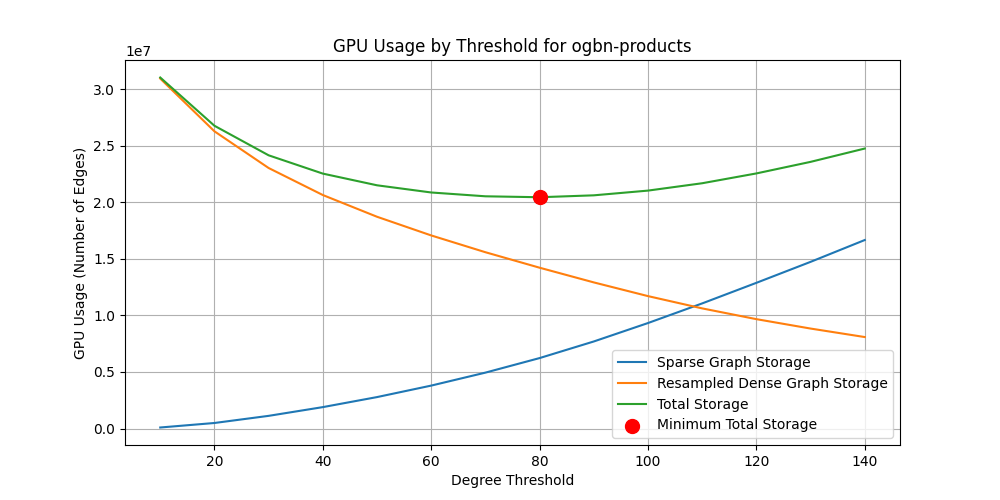}
\caption{GPU Reduction Visualizations for ogbn-products}
\end{figure}

\subsection{Conclusion}

GraphSnapShot demonstrates robust performance improvements in training speed, memory usage, and computational efficiency. By integrating static snapshots with dynamic sampling, GraphSnapShot effectively balances resource utilization and data accuracy, making it an ideal solution for large-scale, dynamic graph learning tasks. Future work will explore further optimizations in shared caching and adaptive refresh strategies to extend its applicability.

% \section{Reference}

% Give out the references

% \bibliographystyle{alpha}
\bibliographystyle{plainnat}
\bibliography{main}

\appendix
\section{Appendix}

\subsection{DGL with GraphSnapShot}

\subsubsection{Datasets}
Table~\ref{tab:ogbn_datasets_overview} summarizes the datasets used in our DGL experiments, highlighting key features like node count, edge count, and classification tasks.

\begin{table}[H]
\centering
\caption{Overview of OGBN Datasets}
\label{tab:ogbn_datasets_overview}
\begin{tabular}{|c|c|c|c|}
\hline
\textbf{Feature} & \textbf{ARXIV} & \textbf{PRODUCTS} & \textbf{MAG} \\ \hline
Type & Citation Net. & Product Net. & Acad. Graph \\ \hline
Nodes & 17,735 & 24,019 & 132,534 \\ \hline
Edges & 116,624 & 123,006 & 1,116,428 \\ \hline
Dim & 128 & 100 & 50 \\ \hline
Classes & 40 & 89 & 112 \\ \hline
Train Nodes & 9,500 & 12,000 & 41,351 \\ \hline
Val. Nodes & 3,500 & 2,000 & 10,000 \\ \hline
Test Nodes & 4,735 & 10,019 & 80,183 \\ \hline
Task & Node Class. & Node Class. & Node Class. \\ \hline
\end{tabular}
\end{table}

\subsubsection{Training Time Acceleration and Memory Reduction}
Tables~\ref{tab:training_time_acceleration} and \ref{tab:runtime_memory_reduction} summarize the training time acceleration and runtime memory reduction achieved by different methods under various experimental settings. 

\begin{table}[H]
\centering
\caption{Training Time Acceleration Across Methods}
\label{tab:training_time_acceleration}
\begin{tabular}{|l|l|c|c|}
\hline
\textbf{Method} & \textbf{Setting} & \textbf{Time (s)} & \textbf{Acceleration (\%)} \\ \hline
FBL             & [20, 20, 20]     & 0.2766                     & - \\ 
                & [10, 10, 10]     & 0.0747                     & - \\ 
                & [5, 5, 5]        & 0.0189                     & - \\ \hline
FCR             & [20, 20, 20]     & 0.2571                     & 7.05 \\ 
                & [10, 10, 10]     & 0.0639                     & 14.48 \\ 
                & [5, 5, 5]        & 0.0163                     & 13.76 \\ \hline
FCR-shared cache& [20, 20, 20]     & 0.2554                     & 7.69 \\ 
                & [10, 10, 10]     & 0.0640                     & 14.33 \\ 
                & [5, 5, 5]        & 0.0161                     & 14.76 \\ \hline
OTF             & [20, 20, 20]     & 0.2460                     & 11.07 \\ 
                & [10, 10, 10]     & 0.0568                     & 23.96 \\ 
                & [5, 5, 5]        & 0.0145                     & 23.28 \\ \hline
OTF-shared cache& [20, 20, 20]     & 0.2393                     & 13.49 \\ 
                & [10, 10, 10]     & 0.0559                     & 25.23 \\ 
                & [5, 5, 5]        & 0.0133                     & 29.63 \\ \hline
\end{tabular}
\end{table}

\begin{table}[H]
\centering
\caption{Runtime Memory Reduction Across Methods}
\label{tab:runtime_memory_reduction}
\begin{tabular}{|l|c|c|c|}
\hline
\textbf{Method} & \textbf{Setting} & \textbf{Runtime Memory (MB)} & \textbf{Reduction (\%)} \\ \hline
FBL & [20, 20, 20] & 6.33 & 0.00 \\ 
    & [10, 10, 10] & 4.70 & 0.00 \\ 
    & [5, 5, 5]    & 4.59 & 0.00 \\ \hline
FCR & [20, 20, 20] & 2.69 & 57.46 \\ 
    & [10, 10, 10] & 2.11 & 55.04 \\ 
    & [5, 5, 5]    & 1.29 & 71.89 \\ \hline
FCR-shared cache & [20, 20, 20] & 4.42 & 30.13 \\ 
                 & [10, 10, 10] & 2.62 & 44.15 \\ 
                 & [5, 5, 5]    & 1.66 & 63.79 \\ \hline
OTF & [20, 20, 20] & 4.13 & 34.80 \\ 
    & [10, 10, 10] & 1.87 & 60.07 \\ 
    & [5, 5, 5]    & 0.32 & 93.02 \\ \hline
OTF-shared cache & [20, 20, 20] & 1.41 & 77.68 \\ 
                 & [10, 10, 10] & 0.86 & 81.58 \\ 
                 & [5, 5, 5]    & 0.67 & 85.29 \\ \hline
\end{tabular}
\end{table}

\subsubsection{GPU Usage Reduction}
GPU memory usage reductions for various datasets are provided in Table~\ref{tab:gpu_usage_reduction}.

\begin{table}[H]
\centering
\caption{GPU Memory Reduction Across Datasets}
\label{tab:gpu_usage_reduction}
\begin{tabular}{|l|c|c|c|}
\hline
\textbf{Dataset} & \textbf{Original (MB)} & \textbf{Optimized (MB)} & \textbf{Reduction (\%)} \\ \hline
OGBN-ARXIV       & 1,166,243             & 552,228                 & 52.65 \\ 
OGBN-PRODUCTS    & 123,718,280           & 20,449,813              & 83.47 \\ 
OGBN-MAG         & 5,416,271             & 556,904                 & 89.72 \\ \hline
\end{tabular}
\end{table}

\begin{figure}[H]
\centering
\begin{minipage}[b]{0.32\textwidth}
    \includegraphics[width=\textwidth]{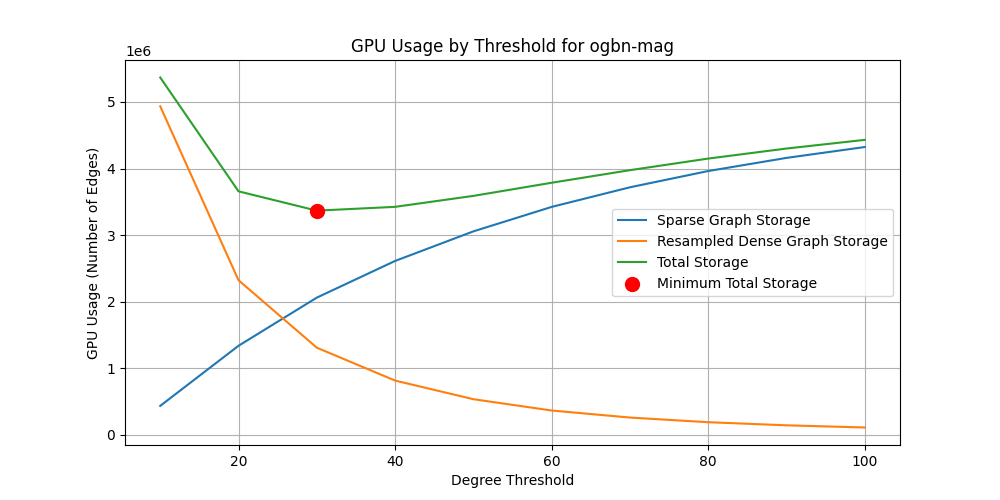}
    \caption{OGBN-MAG GPU Usage}
\end{minipage}
\hfill
\begin{minipage}[b]{0.32\textwidth}
    \includegraphics[width=\textwidth]{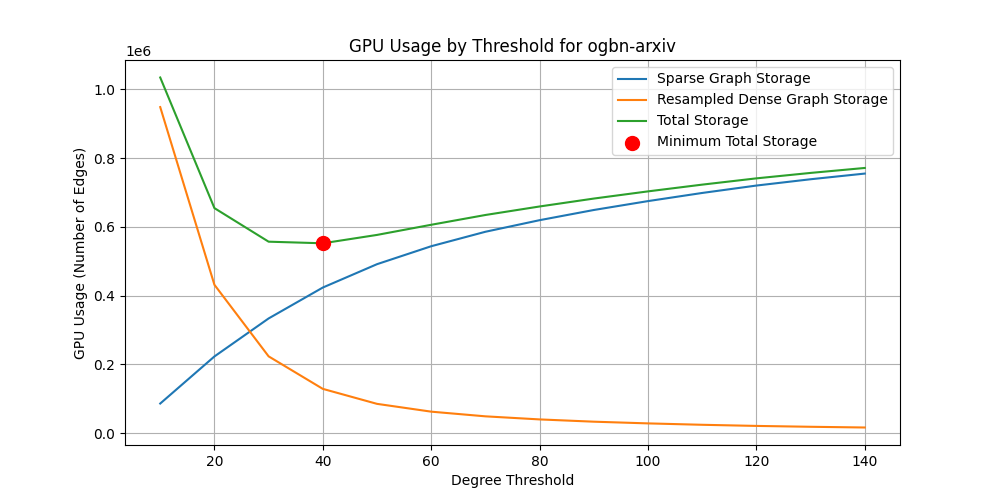}
    \caption{OGBN-ARXIV GPU Usage}
\end{minipage}
\hfill
\begin{minipage}[b]{0.32\textwidth}
    \includegraphics[width=\textwidth]{figures/gpu_by_thrs_products.png}
    \caption{OGBN-PRODUCTS GPU Usage}
\end{minipage}
\end{figure}

\subsection{PyTorch with GraphSnapShot}

% \subsubsection{Datasets}
% We conducted PyTorch experiments on common graph datasets, summarized in Table~\ref{tab:torch_datasets}.

% \begin{table}[H]
% \centering
% \caption{Comparison of PyTorch Experiment Datasets}
% \label{tab:torch_datasets}
% \begin{tabular}{|l|r|r|r|r|}
% \hline
% \textbf{Dataset} & \textbf{Nodes} & \textbf{Edges} & \textbf{Features} & \textbf{Classes} \\ \hline
% PubMed           & 19,717         & 44,338         & 500               & 3                \\ \hline
% Cora             & 2,708          & 5,429          & 1,433             & 7                \\ \hline
% CiteSeer         & 3,312          & 4,732          & 3,703             & 6                \\ \hline
% \end{tabular}
% \end{table}

% \subsubsection{Experimental Setup}
The PyTorch Version GraphSnapShot simulate disk, cache, and memory interactions for graph sampling and computation. Key simulation parameters and operation patterns are listed in Tables~\ref{tab:simulation-duration-access} and \ref{tab:function-access-patterns}.

\begin{table}[H]
\centering
\caption{IOCostOptimizer Functionality Overview}
\label{tab:functionality-overview}
\begin{tabular}{|c|p{8cm}|}
\hline
\textbf{Abbreviation} & \textbf{Description} \\
\hline
Adjust & Adjusts read and write costs based on system load. \\
\hline
Estimate & Estimates query cost based on read and write operations. \\
\hline
Optimize & Optimizes query based on context ('high\_load' or 'low\_cost'). \\
\hline
Modify Load & Modifies query for high load optimization. \\
\hline
Modify Cost & Modifies query for cost efficiency optimization. \\
\hline
Log & Logs an I/O operation for analysis. \\
\hline
Get Log & Returns the log of I/O operations. \\
\hline
\end{tabular}
\end{table}

\begin{table}[H]
\centering
\caption{BufferManager Class Methods}
\label{tab:buffermanager-class-methods}
\begin{tabular}{|p{2cm}|p{8cm}|}
\hline
\textbf{Method} & \textbf{Description} \\
\hline
\texttt{init} & Initialize the buffer manager with capacity. \\
\hline
\texttt{load} & Load data into the buffer. \\
\hline
\texttt{get} & Retrieve data from the buffer. \\
\hline
\texttt{store} & Store data in the buffer. \\
\hline
\end{tabular}
\end{table}

\begin{table}[H]
\centering
\caption{Simulation Durations and Frequencies}
\label{tab:simulation-duration-access}
\begin{tabular}{|c|c|c|}
\hline
\textbf{Operation}           & \textbf{Duration (s)} & \textbf{Simulation Frequency} \\ \hline
Simulated Disk Read          & 5.0011                & 0.05                           \\ \hline
Simulated Disk Write         & 1.0045                & 0.05                           \\ \hline
Simulated Cache Access       & 0.0146                & 0.05                           \\ \hline
In-Memory Computation        & Real Computation      & Real Computation               \\ \hline
\end{tabular}
\end{table}

\begin{table}[H]
\centering
\caption{Function Access Patterns for PyTorch Operations}
\label{tab:function-access-patterns}
\begin{tabular}{|c|c|c|c|}
\hline
\textbf{Operation}           & \textbf{k\_h\_sampling} & \textbf{k\_h\_retrieval} & \textbf{k\_h\_resampling} \\ \hline
Disk Read                    & $\checkmark$           &                          & $\checkmark$              \\ \hline
Disk Write                   & $\checkmark$           &                          & $\checkmark$              \\ \hline
Memory Access                &                        & $\checkmark$             &                           \\ \hline
\end{tabular}
\end{table}

% \subsubsection{Results}
% Tables~\ref{tab:experimental-settings-setting-1} and \ref{tab:experimental-settings-setting-2} summarize the settings used for experiments.

% \begin{table}[H]
% \centering
% \caption{Experimental Settings - Setting 1}
% \label{tab:experimental-settings-setting-1}
% \begin{tabular}{|c|c|c|c|c|}
% \hline
% \textbf{Dataset} & \textbf{Alpha} & \textbf{Presampled} & \textbf{Resampled} & \textbf{Sampled Depth} \\ \hline
% CiteSeer         & 0.1, ..., 0.9 & 100                  & 40                  & 1, 2, 3, 4             \\ \hline
% Cora             & 0.1, ..., 0.9 & 100                  & 40                  & 1, 2, 3, 4             \\ \hline
% PubMed           & 0.1, ..., 0.9 & 100                  & 40                  & 1, 2, 3, 4             \\ \hline
% \end{tabular}
% \end{table}

% \begin{table}[H]
% \centering
% \caption{Experimental Settings - Setting 2}
% \label{tab:experimental-settings-setting-2}
% \begin{tabular}{|c|c|c|c|c|}
% \hline
% \textbf{Dataset} & \textbf{Alpha} & \textbf{Presampled} & \textbf{Resampled} & \textbf{Sampled Depth} \\ \hline
% CiteSeer         & 0.1, ..., 0.9 & 20                   & 10                  & 1, 2, 3, 4             \\ \hline
% Cora             & 0.1, ..., 0.9 & 20                   & 10                  & 1, 2, 3, 4             \\ \hline
% PubMed           & 0.1, ..., 0.9 & 20                   & 10                  & 1, 2, 3, 4             \\ \hline
% \end{tabular}
% \end{table}

\end{document}